\documentclass[12pt]{article}

\usepackage{graphicx} 
\usepackage{hyperref}
\usepackage{times}
\usepackage{geometry}
\usepackage{fancyhdr}
\usepackage{tabularray}
\usepackage{float}
\usepackage{geometry}
\usepackage{tabularx}
\usepackage{etoolbox} 

\title{Easy Problems That\\ LLMs Get Wrong}

\author{
  Sean Williams*\\
  \texttt{sean@autogenai.com}
  \and
  James Huckle*\\
  \texttt{james@autogenai.com}
}
\date{May 2024}

\begin{document}
\enlargethispage{\baselineskip}
\enlargethispage{\baselineskip}
\enlargethispage{\baselineskip}
\enlargethispage{\baselineskip}
\setlength{\parindent}{0pt}
\maketitle

\begin{center}
\large \textbf{Abstract}
\end{center}

\begin{quote}
We introduce a comprehensive Linguistic Benchmark designed to evaluate the limitations of Large Language Models (LLMs) in domains such as logical reasoning, spatial intelligence, and linguistic understanding, among others. Through a series of straightforward questions, it uncovers the significant limitations of well-regarded models to perform tasks that humans manage with ease. It also highlights the potential of prompt engineering to mitigate some errors and underscores the necessity for better training methodologies. Our findings stress the importance of grounding LLMs with human reasoning and common sense, emphasising the need for human-in-the-loop for enterprise applications. We hope this work paves the way for future research to enhance the usefulness and reliability of new models.
\end{quote}

\section{Introduction}
Large Language Models (LLMs) have emerged as a transformative technology with utility across a range of applications. Despite their impressive capabilities, LLMs exhibit notable deficiencies that hinder their ability to comprehend and reason robustly, raising questions about the breadth and depth of their isolated applicability without human oversight.

\

In response to these challenges, we propose a Linguistic Benchmark comprising of 30 questions designed to asses their well-documented limitations across domains such as spatial reasoning, linguistic understanding, relational thinking, mathematical reasoning, and knowledge of basic scientific concepts. This benchmark is not merely an academic exercise but a tool to gauge the current capabilities of LLMs in areas they are well-documented to fail.

\

Furthermore, we explore an often-underestimated facet of interacting with these models: prompt engineering. By refining the manner in which tasks are presented to LLMs, we can significantly influence their output, guiding them towards more accurate and logically sound responses. Our work underscores the need for continued innovation in the field, extending what these extraordinary models can already achieve while conscientiously addressing their limitations.

\

The code relevant to this paper can be found at our \href{https://github.com/autogenai/easy-problems-that-llms-get-wrong}{\textbf{GitHub repository}}.

\rule{\linewidth}{0.4pt}\par
\noindent * Authors contributed equally

\section{Background and Related Work}
\subsection{Known LLM Limitations}
\subsubsection{Linguistic Understanding}
LLMs have demonstrated difficulties with linguistic understanding, stemming primarily from the limitations of their intrinsic operational mechanisms. These models often misinterpret or overlook the nuanced meanings conveyed in human language. This results in inaccuracies or misjudgments when dealing with linguistic tasks or when parsing sentences that demand a deeper grasp of idiomatic nuances. \cite{Linguistic Understanding}

\subsubsection{Common Sense}
A pivotal hindrance of LLMs lies in their absence of embodied experience, a factor Philosopher Hubert Dreyfus highlighted as crucial for the development of common sense in humans \cite{Common Sense}. Unlike humans, who engage with the physical world through a rich palette of sensory experiences such as visual, auditory, and tactile stimuli, LLMs operate without sensory perception. This disembodied state restricts their capacity to learn the subtleties inherent to commonsense reasoning \cite{Common Sense 2}.

\subsubsection{Contextual Understanding}
AI's inability to handle context-sensitive reasoning was also critiqued by Dreyfus, which remains relevant to today's LLMs \cite {Contextual Understanding}. Correct reasoning is deeply intertwined with the ability to understand the often implicit context in which something relates.

\subsubsection{Visual-Spatial Reasoning}
Visual-spatial reasoning entails the ability to mentally visualise objects and understand their relationship within space. LLMs lack fundamental spatial awareness, so explaining the steps needed to navigate from one point to another in physical space or understanding the spatial configuration of objects remains a complex challenge for these models, showcasing a significant gap in a vital area of human intelligence.  \cite {Spatial Understanding}

\subsubsection{Mathematical Reasoning}
LLMs express fragility in conducting simple mathematical reasoning, with word-based tasks that involve counting to ten often posing a significant challenge. While they can often provide correct answers to sophisticated mathematical queries, they fundamentally lack a rules-based counting system. They must outsource their calculations to other tooling, such as computer code or a calculator.  \cite{Counting}

\subsubsection{Popular Science Knowledge}
LLMs are particularly vulnerable to propagating and reinforcing inaccuracies found within their training data, including scientific misconceptions or outdated information commonly perpetuated online. An LLM's approach to generating content is heavily reliant on the frequency and presentation of information encountered during their training, which can lead to the uncritical replication of errors. This propensity not only highlights the limitations in their comprehension abilities but also underscores the importance of curating and updating the data these models are trained on to mitigate the dissemination of incorrect or misleading information. \cite{Science}

\subsubsection{Relational Understanding}
Understanding and interpreting relationships between entities, whether temporal, causal, or conceptual, is another area where LLMs face difficulties. Their interpretations often lack the depth and nuance of human understanding, as solving relational problems often requires an intuition that LLMs inherently lack. \cite{Relational}

\subsubsection{Logical Reasoning}
LLMs are trained on an encompassing array of knowledge; however, this training approach does not guarantee proficiency in logical reasoning at inference time. Previous research has explored the logical capabilities of LLMs with mixed findings and suggests that LLMs can mimic reasoning up to a certain level but lack the reliability of human-like reasoning \cite{Logical}.

\subsubsection{Overfitting}
Overfitting is a well-documented phenomenon in machine learning, where models excessively adapt to the idiosyncrasies of the training data at the expense of broader generalisation. It is the belief that pre-trained models should excel in interpolating within the bounds of their training data but that extrapolation outside of those bounds is more difficult  \cite{Overfitting}.

\subsection{Popular Benchmark Obsession}
Various technology providers have developed their own LLMs, leading to a competitive ecosystem of models. These models are typically evaluated using standard benchmarks, and any advancements receive considerable media attention. There is a concern that these popular benchmarks, even those involving independent human evaluation, may encourage a myopic focus on "ranking" at the expense of optimising for holistic performance. This is a plausible manifestation of Goodhart's maxim, commonly stated as - ``When a measure becomes a target, it ceases to be a good measure."

\subsubsection{Performance Gap}
Hands-on experience with different models often exposes a larger performance gap, both between the models and in absolute terms, than is indicated by popular benchmarks, highlighting the possible inadequacies of these tests \cite{benchmark failure}. Novel benchmarks that can expose the truer magnitude of their difference in performance by quizzing them with difficult tasks could be a valuable asset to the AI community and to commercial enterprises that increasingly rely on these models.

\newpage
\section{Methodology}
This section presents a collection of questions developed to be easy for human adults to answer but challenging for LLMs. These questions serve as a linguistic benchmark to examine model performance in several key domains where they have known limitations. This benchmark is useful for monitoring the performance of LLMs over time and highlighting their failure modes.

\subsection{Linguistic Benchmark}
\textbf{All benchmark questions can be found in the Appendix} \ref{app:Benchmark_questions}.

\subsubsection{Question Taxonomy}
\begin{table}[ht!]
\centering
\begin{tabular}{|>{\raggedright\arraybackslash}p{0.19\linewidth}|>{\raggedright\arraybackslash}p{0.75\linewidth}|}
\hline
\textbf{Question Type} & \textbf{Description} \\
\hline
Puzzle & Logic-type puzzles that mimic the structure of popular questions found online but differ significantly in one or more critical aspects that make the questions much easier for humans. \\
\hline
Spatial & Requires visualising the arrangement or relative positions of objects in space, such as determining the order or position of items or simple navigation. \\
\hline
Relational & Involve understanding and inferring relationships or hierarchies between objects, concepts, or entities based on provided information.  \\
\hline
Counting & Simple numerical calculations such as counting to a maximum of ten or understanding quantities. \\
\hline
Linguistic & Tests the understanding and use of language, including forming sentences with specific constraints or identifying unique characteristics of words and phrases. \\
\hline
Popular science & Straightforward questions that test for common scientific and mathematical misconceptions  \\
\hline
\end{tabular}
\caption{Linguistic Benchmark Question Types}
\end{table}

\subsubsection{Benchmark Inclusion Qualification}
Inclusion in the Linguistic Benchmark mandates that a question must be deemed easy for the average adult to answer, align with one of the specified taxonomic classifications, and, upon evaluation, result in at least one of the designated LLMs failing to produce a correct response. Rather than establishing a formal definition of``easy," this criterion was applied based on a best-efforts basis.

\newpage
\subsection{LLM Selection and Hyperparameters}
For our research, we selected an array of popular Large Language Models (LLMs), encompassing offerings from industry leaders such as OpenAI, Anthropic, Mistral, Google, and Meta. This assortment comprises both proprietary and open-source models.

\

\begin{table}[H]
\centering
\begin{tabular}{l l l}
\hline
\textbf{Provider} & \textbf{Model Version} & \textbf{Inference Type} \\
\hline
OpenAI & GPT-4 Turbo Preview & OpenAI API \\
& \textit{gpt-4-0125-preview} & \\
Anthropic & Claude 3 Opus & Anthropic API \\
& \textit{claude-3-opus-20240229} & \\
Mistral & Mistral Large & Mistral API \\
& \textit{mistral-large-latest} & \\
Mistral & Mistral 8x22B & Mistral API \\
& \textit{open-mixtral-8x22b} & \\
Google & Gemini - Pro 1.0 & GCP Vertex API \\
& \textit{gemini-1.0-pro-002} & \\
Google & Gemini - Pro 1.5 & GCP Vertex API \\
& \textit{gemini-1.5-pro-001} & \\
Meta & Llama 3 70B & AWS Bedrock API \\
& \textit{Llama 3 70B Instruct v1-ODT} & \\
\hline
\end{tabular}
\caption{Comparison of language model settings}
\label{table:language_model_settings}
\end{table}

\subsubsection{Inference Type}
We queried most of the models via the API offered directly by the providers to achieve a stable and unaltered performance, with the exception of Meta Llama 3 70B, which was queried using AWS Bedrock. To the best of our knowledge, none of the models were quantised, so we will assume they were running at full precision.

\subsubsection{Hyperparameters}
All of the model hyperparameters were set to their default values at the time of testing, with the exception of the `temperature`, which was set to zero where it was present. A `temperature` of zero is preferred because the output becomes mostly deterministic (see Appendix \ref{app:LLM_determinism} for more details), aiding study reliability and reproducibility. Additionally, any absolute positive value for 'temperature' may be represented differently across each LLM architecture, some of which are closed-source and cannot be fully known. 

\

The authors understand that higher temperatures would allow for a wider variety of answers and, therefore, alter the probability that a model could produce a more accurate answer due to a more thorough discovery of the token distribution space (this is addressed further in Section  \ref{app:Future Work and Limitations}, ``Future Work and Limitations"). However, some studies suggest that increasing the temperature "does not have a statistically significant impact on LLM performance for problem-solving tasks" \cite{LLM_temperature}.

\newpage
\subsection{Evaluation Process}
Models underwent evaluation against a structured scoring framework for each question within the Linguistic Benchmark. This framework is designed to assess the precision of answers, accuracy of reasoning, and conformity to logical principles. The evaluation process was conducted from April 14th to April 28th, 2024.

\subsubsection{Scoring Criteria}
The authors employed a predefined scoring system, ranging from 0\% to 100\%, to evaluate the responses. This assessment framework is designed to gauge various aspects of the responses, including: the accuracy of the answer provided, the soundness of the reasoning process, the relevance and usefulness of the information given, and the presence of any logical discrepancies. The system aims to deduct points from responses that present both correct and incorrect answers, which impacts their credibility.

\begin{table}[H]
\centering
\begin{tabular}{| c | p{13cm} |}
\hline
\textbf{Score} & \textbf{Marking Criteria}                                                                                                             \\ \hline
100\% & The response contains the correct answer only with a correct thought process and no logical inconsistencies. \\ \hline
80\%  & The response contains the correct answer only with a correct thought process with some logical inconsistencies. \\ \hline
60\% & The response contains the correct answer only but with an incorrect thought process. \\ \hline
40\% & The response contains an incorrect answer anywhere but also provides a correct answer or correct thought process with minimal logical inconsistencies. \\ \hline
20\% & The response contains an incorrect answer anywhere but provides enough helpful information to plausibly reach a correct answer. \\ \hline
0\%  & The response contains an incorrect answer, too much unhelpful information, or not enough helpful information to plausibly reach a correct answer. \\ \hline
\end{tabular}
\caption{Scoring Rubric}
\label{tab:grading_rubric}
\end{table}

\subsubsection{Scoring Methods}
The primary evaluation method involved bind manual scoring of each question by the authors. This methodology was chosen because whilst using an LLM for scoring open-ended responses offers convenience, it frequently lacks the necessary rigour and reliability. Nonetheless, automated evaluations were attempted and are presented as supplementary material. The code and prompt templates used for this process can be found in the paper's \href{https://github.com/autogenai/easy-problems-that-llms-get-wrong}{GitHub repository} for further reference and examination.

\subsubsection{Human Benchmark}
As of now, 14 individuals have completed our quiz and have thus contributed to the human benchmark score of our Linguistic Benchmark, achieving an average score of 86\%. We would be honoured to have you as part of this study, so if you are interested, we kindly \textbf{invite you to take our 
\href{https://forms.office.com/r/vND5bfiFHd}{Online Quiz}}

\newpage
\section{Results}
\subsection{LLM Responses}

To maintain brevity, we have presented the responses from each provider's top-performing model for the first ten out of the thirty questions in our benchmark. We also provide a typical correct human response for comparison. \textbf{Responses can be found in the Appendix }\ref{app:LLM_responses}

\

The full set of responses can be found at our \href{https://github.com/autogenai/easy-problems-that-llms-get-wrong}{\textbf{GitHub repository}}.

\subsection{Table of Results}
The results were aggregated, with scores averaged across all questions. To assess the reliability of these findings, bootstrapped confidence interval analysis was performed with a sample size of 10,000, operating at a 95\% confidence level.

\begin{table}[H]
\centering
\begin{tabular}{l c c}
\hline
\textbf{Model} & \textbf{Average Score} & \textbf{CI (95\%)} \\
\hline
GPT-4 Turbo & 38\% & [23\% , 55\%] \\
Claude 3 Opus & 35\% & [21\% , 52\%] \\
Gemini 1.5 Pro & 30\% & [15\% , 45\%] \\
Mistral Large & 28\% & [15\% , 42\%] \\
Llama 3 70B & 27\% & [14\% , 43\%] \\
Mistral 8x22B & 20\% & [7\% , 34\%] \\
Gemini 1.0 Pro & 16\% & [6\% , 29\%] \\
\hline
\end{tabular}
\caption{Performance on Linguistic Benchmark (30)}
\label{table:logic problem dataset performance}
\end{table}

\begin{figure}[H]
    \centering
    \includegraphics[width=1\linewidth]{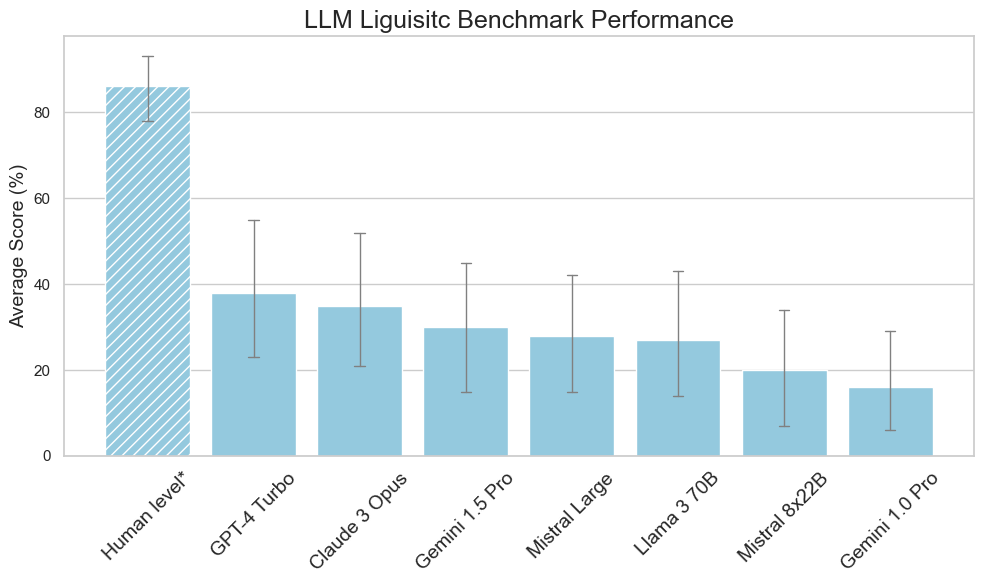}
    \caption{LLM Linguistic Benchmark Performance}
    \label{fig:enter-label}
\end{figure}

\subsubsection{Automated Scoring Results}
\textbf{See Appendix \ref{app:Automated Scoring}}

\subsection{Clarifying Questions Improved Performance}
\label{app:Clarifying Questions Section}
Subsequent experiments involved a multi-step process starting with the models requesting clarifying questions to enhance their comprehension of the original queries. Once these clarifying questions were addressed with relevant answers, the models were asked to respond to the initial questions.

\

\textbf{An example can be found in the Appendix }\ref{app:Clarifying_questions}.

\subsubsection{Improvement in Results}
Asking the model to provide clarification questions resulted in a relative improvement of 40.7\% across the models tested. In total, 71 answers demonstrated improvements, 128 answers remained unchanged, and 16 answers suffered regressions.

\begin{table}[H]
\centering
\begin{tabular}{l c r r r}
\hline
 & \textbf{Average Score} & \multicolumn{3}{c}{\textbf{Change from Baseline}} \\
\cline{3-5}
 & & \textbf{Absolute} & \multicolumn{2}{c}{\textbf{Relative}} \\
\hline
GPT-4 Turbo & 52\% & +14\% & \hspace{1em} +36.8\% \\
Claude 3 Opus & 48\% & +13\% & \hspace{1em} +37.1\% \\
Llama 3 70B & 47\% & +20\% & \hspace{1em} +74.1\% \\
Mistral Large & 39\% & +11\% & \hspace{1em} +39.3\% \\
Mistral 8x22B & 38\%& +18\% & \hspace{1em} +90.0\% \\
Gemini 1.5 Pro & 32\% & +2\% & \hspace{1em} +6.7\% \\
Gemini 1.0 Pro & 17\% & +1\% & \hspace{1em} +6.3\% \\
\hline
\end{tabular}
\caption{Performance after clarifying questions}
\label{tab:clarifying-questions-performance}
\end{table}

\section{Discussion}
Our research reveals that Large Language Models (LLMs) often struggle to answer questions that adult humans find straightforward and intuitive. This fact poses significant challenges for their use in areas that demand strong logical reasoning skills and operation without human supervision. Our results indicate that the present training methods do not sufficiently equip LLMs for tasks requiring novel forms of reasoning. Moreover, it's hard to predict how and when they might fail at such tasks.

\subsection{Common LLM Failings}

\subsubsection{Overfitting} 
The models frequently yielded incorrect, convoluted, and verbose responses to our logic-based puzzles. They would often default to solutions corresponding to the original versions of these puzzles found online instead of correctly addressing our benchmark questions - which are deliberately modified versions. This behaviour underscores a tendency towards LLMs overfitting their web-based training corpus, adversely impacting their proficiency to generate accurate responses to novel problems (e.g. Question 2, Claude 3 Opus, Appendix \ref{app:Question 2}).

    \begin{quote}
    \textit{Question:} Suppose you’re on a game show, and you’re given the choice of three
    doors: Behind one door is a gold bar; behind the others, rotten
    vegetables. You pick a door, say No.1, and the host asks you “Do you
    want to pick door No.2 instead?” Is it to your advantage to switch
    your choice?

    \textit{Claude 3 Opus Answer:} ``This is a classic probability problem known as the “Monty Hall problem.” The answer is that it is always to your advantage to switch your choice. Here’s why: ..."
    \end{quote}

\subsubsection{Lack of Logic or Common Sense}
There were examples of illogical statements and commonsense inconsistencies within a sentence (e.g. Question 4, Gemini 1.0 Pro, Appendix \ref{app:Question 4}).

    \begin{quote}
    ``... Now the farmer can safely leave the wolf with the goat because the cabbage is no longer a threat. ..."
    \end{quote}

\subsubsection{Lack of Spatial Intelligence}
Problems involving simple visual-spatial skills, such as those that required understanding the movement of a revolving cylinder or an arrangement of people, caused problems (e.g. Question 22, GPT-4 Turbo, \href{https://github.com/autogenai/easy-problems-that-llms-get-wrong/blob/84570900736ec958752badf19502210f0245c031/llm_outputs/final_answers-gpt-4-turbo.json\#L149}{GitHub repository}).

    \begin{quote}
    \textit{Question:} I’m in London and facing west, is Edinburgh to my left or my right?

    \textit{GPT-4 Turbo Answer:} ``If you're in London facing west, Edinburgh would be to your left, as it is located to the north of London."
    \end{quote}

\subsubsection{Incorrect Mathematical Reasoning}
Some answers displayed a fundamental misunderstanding of counting principles and performed arithmetic errors that led to incorrect conclusions. This failure mode was identified in problems that require accurate numerical manipulation or enumeration (e.g. Question 23, GPT-4 Turbo, \href{https://github.com/autogenai/easy-problems-that-llms-get-wrong/blob/84570900736ec958752badf19502210f0245c031/llm_outputs/final_answers-gpt-4-turbo.json\#L156}{GitHub repository}).

    \begin{quote}
    \textit{Question:} Count the number of occurrences of the letter 'L' in the word -
    
    'LOLLAPALOOZA'.
    
    \textit{GPT-4 Turbo Answer:} ``... There are 5 'L's in the word 'LOLLAPALOOZA'"
    \end{quote}

\subsubsection{Poor Linguistic Understanding}
Challenges in understanding nuances and interpretations of simple linguistic tasks were evident. This showcased a limitation in LLMs' processing of linguistic constraints, reflecting a gap in their understanding of language at depth (e.g. Question 8, Mistral Large, Appendix \ref{app:Question 8}).

    \begin{quote}
    \textit{Question:} Write me a sentence without any words that appear in The Bible.
    
    \textit{Mistral Large Answer:} ``Intricate quantum particles danced in the nebulous cosmos, transcending conventional spatiotemporal dimensions."
    \end{quote}

\subsubsection{Popular Science}
The LLMs struggled with appropriately addressing questions common in popular science, failing to apply scientific knowledge accurately or misunderstanding the fundamental principles involved. This issue becomes particularly evident in tasks that demand an intuitive grasp of physical laws, chemical properties, or biological processes (e.g. Question 9, GPT-4 Turbo, Appendix \ref{app:Question 9}).

    \begin{quote}
    \textit{Question:} Which weighs more, a pound of water, two pounds of bricks, a pound of feathers, or three pounds of air?
    
    \textit{GPT-4 Turbo Answer:} ``Two pounds of bricks weigh more than a pound of water, a pound of feathers, or three pounds of air. ..."
    \end{quote}

\subsubsection{Relational Misunderstandings}
LLMs often fail to understand or process relational contexts correctly, leading to incorrect interpretations and conclusions about the relationships between entities or concepts. Tasks that required an understanding of hierarchical relationships or include cause-and-effect dynamics were especially troublesome (e.g. Question 21, GPT-4 Turbo, \href{https://github.com/autogenai/easy-problems-that-llms-get-wrong/blob/84570900736ec958752badf19502210f0245c031/llm_outputs/final_answers-gpt-4-turbo.json\#L142}{GitHub repository}).

    \begin{quote}
    \textit{Question:} Sally (a girl) has 3 brothers. Each brother has 2 sisters. How many sisters does Sally have?
    
    \textit{GPT-4 Turbo Answer:} ``Sally has 2 sisters. Each of her brothers has 2 sisters: Sally and another sister, making a total of 3 girls in the family."
    \end{quote}

\subsubsection{Illogical Chain of Thought}   
There were examples of logical inconsistencies and falsehoods within the Chain of Thought (CoT) process, contradicting the final answer. This highlights the fragility of CoT in adversarial settings. (e.g. Question 3, Claude 3 Opus, Appendix \ref{app:Question 3}).

    \begin{quote}   
    ``From a probability standpoint, it does not matter whether he spins the chambers again or not. Here’s why:
    
    Scenario 1 (No spin): ... The probability ... is 1 (or 100\%).
    
    Scenario 2 (Spin): ...  The probability ... is 5/6 (or about 83.33\%)
    
    In both scenarios, the expected outcome is the same: a 5/6 chance ..."
    \end{quote}

\subsection{LLM Performance} 
\subsubsection{Linguistic Benchmark} 
There was a noticeable variance in the performance across popular LLMs, with OpenAI's GPT-4 Turbo demonstrating markedly more robust logical reasoning throughout our tests. Perhaps unsurprisingly, the smaller and open-source models Llama 3 70B and Mistral 8x22B had the lowest benchmark performance aside from the previous generation, Gemini 1.0 Pro. However, both open-source models demonstrated a strong performance improvement after introducing clarifying questions, highlighting their increased capacity for optimisation through prompt engineering.

\subsubsection{Popular LLM Benchmarks}
In the rapidly evolving world of AI, benchmarks serve a crucial function in assessing the performance of new models. These benchmarks are sets of questions designed to test various facets of a language model's ability. Comprehension, mathematics, and general knowledge are the most significant areas assessed.

\

\begin{table}[H]
\begin{tabular}{lcccccc}
\hline
\textbf{Model} & \textbf{MMLU} & \textbf{GSM8K} & \textbf{BIG-Bench H} & \textbf{DROP} & \textbf{HellaSwag} & \textbf{ARC C} \\ \hline
\textit{\textbf{Claude 3 Opus}} & \textit{\textbf{86.8}}\% & \textit{\textbf{95.0}}\% & \textbf{\textit{86.8}}\% & \textit{\textbf{83.1}}\% & \textit{\textbf{95.4}}\% & \textit{\textbf{96.4}}\% \\
GPT-4 Turbo & 86.4\% & 92.0\% & 83.1\% & 80.9\% & 95.3\% & 96.3\% \\
Llama 3 70B & 82.0\% & 93.0\% & --- & --- & --- & --- \\ 
Gemini 1.5 Pro & 81.9\% & 91.7\% & 84.0\% & 78.9\% & 92.5\% & --- \\ 
Mistral Large & 81.2\% & 91.2\% & --- & --- & 89.2\% & 94.2\% \\
Mistral 8x22B & 77.8\% & 76.5\% & --- & --- & 88.6\% & 91.3\% \\
Gemini 1.0 Pro & 71.8\% & 86.5\% & 75.0\% & 74.1\% & 84.7\% & --- \\ 
\hline
\end{tabular}
\caption{Popular Benchmarks}
\label{table: performance on published benchmarks}
\end{table}

References: \href{https://www.anthropic.com/news/claude-3-family}{\textit{Anthropic}},
\href{https://openai.com/index/gpt-4-research/}{\textit{OpenAI}},
\href{https://llama.meta.com/llama3/}{\textit{Meta}},
\href{https://deepmind.google/technologies/gemini/pro/}{\textit{Google}},
\href{https://mistral.ai/news/mixtral-8x22b/}{\textit{Mistral}}.

\subsubsection{The Pressure to Perform}
The outcomes of well-recognised benchmarks are widely promoted and emphasised in the industry. When a new model surpasses the incumbent on any of these benchmarks, it is viewed as a major advancement and gains substantial attention. One such example is Claude 3 Opus, which demonstrated exceptional performance against GPT-4 Turbo across many standard benchmarks; however, it did not perform as strongly on our specialised Linguistic Benchmark. This highlights a potential downside for companies focusing too intensely on excelling in known benchmarks: they risk developing models that excel in benchmark scenarios but may not adapt as effectively to a broader range of challenges or unforeseen tasks. \cite{LLM_benchmark_cheater}

\subsubsection{Novel Human-Level Benchmarks}
Despite the proliferation of benchmarks, there remains a gap between an LLM's theoretical accomplishments and its real-world effectiveness, especially in comparison with human-level performance on novel tasks. Current benchmarks, while diverse in nature, tend to focus on narrow domains or specific types of reasoning, failing to capture the nuance of regular tasks. 

\subsubsection{Test Set Leakage}
It is plausible that our own Linguistic Benchmark may one day be used to train a foundational LLM model, degrading its utility as a novel test set. It may be possible to verify this by confirming its inclusion in any common pre-training datasets or by monitoring the change in log probabilities of token outputs from partially inserted questions. 

\subsection{LLM Reliability}
\subsubsection{Input Structure}
Minute changes to input structure or order that do not change the meaning of a question lead to dramatically different responses from LLMs. Section \ref{app:Clarifying Questions Section}    on ``Clarifying Questions Improved Performance" highlighted this weakness; a handful of previously correct answers suffered regressions across the models tested despite the underlying question remaining the same.

\subsubsection{Output Determinism}
A frequently reported but largely undocumented phenomenon is that of non-deterministic LLM outputs with the temperature set to 0. We noticed this non-deterministic behaviour in all models and providers, which makes it increasingly difficult to perform repeatable benchmarking and complicates the development of prompt engineering. It also highlights the real need for continuous testing. \\

\textbf{More details in the Appendix} \ref{app:LLM_determinism}

\subsection{LLMs Recommend Adversarial Examples}
When prompted, GPT-4 Turbo is apt at recommending other adversarial logic-type puzzles that may cause it to provide an ``overfitted response". This is how the authors found ``Question 2," which is loosely structured linguistically on the famous Monty Hall Problem.

\

\textbf{Example in the Appendix} \ref{app:Adversarial_Examples}

\subsubsection{Inclusion of LLM Recommendations in the Linguistic Benchmark}
The authors feel it is unproblematic to include questions that were provided by a single LLM to test other models and, by extension, do not believe this imparts any appreciable bias. This is due to the datasets used to train nearly all general-purpose LLMs being highly correlated. Common Crawl, WebText2, Books1, Books2, Wikipedia, and news articles constitute the vast majority of trained tokens \cite{liu2024datasets}. The paper ``Language Modeling Is Compression" published by Google DeepMind \cite{DeepMindCompression} suggests that LLMs learn a compressed representation of the data they are trained on. Therefore, it is unsurprising that all of the models tested provided similar adversarial logic-type problem recommendations. 

\newpage
\section{Implications}
The findings from this benchmark highlight several key implications for the development, deployment, and expectation management of Large Language Models (LLMs):

\subsection{Quality Over Quantity}
The inconsistencies and specific types of failures indicated in the benchmark suggest that future development of LLMs should prioritise not only scale but also the quality of reasoning and reliability across a wider array of questions. This includes improving logical reasoning, spatial intelligence, linguistic understanding, and commonsense reasoning. Moreover, the ability to understand and process relational contexts accurately should be enhanced. Adopting diverse and comprehensive datasets with an emphasis on curating adversarial or challenging problems during training might help in addressing some of these shortcomings.

\subsection{Commercial Use}
Organisations planning to deploy LLMs must be cautious about relying on them for tasks that require high-stakes decision-making, nuanced reasoning, or understanding subtle linguistic cues unless supervised or complemented by human judgment. Deployment strategies should be adaptable, incorporating continuous monitoring for failure modes, regular benchmarking against novel problem sets, and readiness to integrate human oversight whenever the models' limitations are encountered.

\subsection{Addressing Overfitting and Benchmark Limitations}
While benchmarks are useful for standardised evaluations, they should be complemented alongside more dynamic and unpredictable tests reflecting real-world complexity.

\subsection{Promoting Openness and Collaboration}
Sharing findings, particularly regarding failure modes, can foster a collective effort toward addressing these limitations. Such collaboration might not only accelerate individual efforts but could also lead to the development of more versatile and reliable AI systems.

\subsection{Acknowledging Limitations}
The limitations and unpredictability in LLM performance observed underscore the importance of responsible development. Model developers and deploying organisations must be transparent about the capabilities and limitations of their systems, ensuring users are informed of the potential for error or bias. The development of LLM models should include rigorous testing to uncover and address potential failure modes before widespread deployment.

\subsection{Enhancing Input and Output Stability}
Our findings indicate a need for LLMs to improve the handling of subtle variations in input and ensure consistent, reliable outputs. Additionally, offering the ability to guarantee a deterministic output would be helpful for many use cases.

\subsection{Research Direction}
Finally, this benchmark opens new avenues for research, particularly in exploring methods to improve LLMs’ linguistic understanding and comprehension. It also raises questions about how these models conceptualise and process different forms of logic and common sense. Enhancing model performance will likely require an interdisciplinary approach that blends cognitive science, linguistics, and artificial intelligence research.

\section{Future Work and Limitations}
\label{app:Future Work and Limitations}
There are vast limitations to this approach, but further improvements might include:
\begin{itemize}
    \item Expanding the Linguistic Benchmark beyond thirty questions to increase statistical significance and test a more diverse range of inputs.
    \item Using multiple-choice questions to make evaluation more reliable.
    \item Running inference multiple times with the temperature for each model set above zero (standardised and equivalent across all architectures) and generating aggregate statistics. 
    \item Testing on a sample of smaller LLMs to see if performance is correlated to model size.
    \item Fine-tuning models with a training dataset of perturbed variations of well-known logic-type problems found in the training corpora (on the internet) to see if this decreases overfitting variance.
    \item Testing advanced regularisation techniques for LLMs during the pre-training process.
    \item Finding better methodologies to keep LLM outputs deterministic.
\end{itemize}

\section{Conclusion}
The Linguistic Benchmark suggests that LLMs offered by leading providers such as OpenAI, Anthropic, Meta, Mistral, and Google have difficulty answering novel questions that humans find relatively easy. These models falter across domains such as logical reasoning, spatial intelligence, mathematical reasoning, linguistic understanding, knowledge of popular science, and relational perception. This highlights a significant gap between their current performance and general human cognitive abilities. Spotlighting areas where LLMs underperform invites a re-calibration of our expectations for these models, encouraging a focus on enhancing their reasoning capabilities and a pivot towards human-in-the-loop augmented intelligence.

\

This research reminds us of the importance of responsible LLM deployment. As we integrate LLMs into various facets of societal operations, from education to healthcare, acknowledging and addressing their limitations is paramount. This not only ensures that we leverage these models' strengths but also safeguards against potential risks and pitfalls. Whilst significant strides have been made in the development of LLMs, this investigation underscores the significant challenges that remain in developing models that can achieve human-like understanding and reasoning. The Linguistic Benchmark offers a critical perspective, urging LLM development that prioritises higher standards of general purpose reliability.

\newpage
\section{Appendix}

\subsection{Linguistic Benchmark Questions}
\label{app:Benchmark_questions}

\begin{table}[ht!]
\centering
\begin{tabular}{|>{\raggedright\arraybackslash}p{0.05\linewidth}|>{\raggedright\arraybackslash}p{0.15\linewidth}|>{\raggedright\arraybackslash}p{0.75\linewidth}|}
\hline
\textbf{No.} & \textbf{Category} & \textbf{Question} \\
\hline
1 & Puzzle & You have six horses and want to race them to see which is fastest. What is the best way to do this? \ \ref{app:Question 1} \\
\hline
2 & Puzzle & Suppose you're on a game show, and you're given the choice of three doors: Behind one door is a gold bar; behind the others, rotten vegetables. You pick a door, say No. 1, and the host asks you ``Do you want to pick door No. 2 instead?" Is it to your advantage to switch your choice? \ \ref{app:Question 2} \\
\hline
3 & Spatial & You are playing Russian roulette with a six-shooter revolver. Your opponent puts in five bullets, spins the chambers and fires at himself, but no bullet comes out. He gives you the choice of whether or not he should spin the chambers again before firing at you. Should he spin again? \ \ref{app:Question 3} \\
\hline
4 & Puzzle & A farmer wants to cross a river and take with him a wolf, a goat and a cabbage. He has a boat with three secure separate compartments. If the wolf and the goat are alone on one shore, the wolf will eat the goat. If the goat and the cabbage are alone on the shore, the goat will eat the cabbage. How can the farmer efficiently bring the wolf, the goat and the cabbage across the river without anything being eaten?  \ \ref{app:Question 4}\\
\hline
5 & Puzzle & Bob has three boxes in front of him - Box A, Box B and Box C. Bob does not know what is in the boxes. Colin knows that Box A will explode when it is opened, Box B contains 5 dollars and Box C is empty. Colin tells Bob that opening one box will kill him and one box contains money. Should Bob open a box? \ \ref{app:Question 5} \\
\hline
6 & Counting & A robot has 8 arms. There are 5 objects on a table: a knife, a fork, a spoon, a teddy bear and a doll. The robot picks up each object with an arm. He then shakes hands with himself. \ \ref{app:Question 6} \\
\hline
7 & Spatial & Alan, Bob, Colin, Dave and Emily are standing in a circle. Alan is on Bob’s immediate left. Bob is on Colin’s immediate left. Colin is on Dave’s immediate left. Dave is on Emily’s immediate left. Who is on Alan’s immediate right? \ \ref{app:Question 7} \\
\hline
8 & Linguistic & Write me a sentence without any words that appear in The Bible. \ \ref{app:Question 8} \\
\hline
9 & Popular science & Which weighs more, a pound of water, two pounds of bricks, a pound of feathers, or three pounds of air. \ \ref{app:Question 9} \\
\hline
10 & Relational & I get out on the top floor (third floor) at street level. How many stories is the building above the ground? \ \ref{app:Question 10} \\
\hline
\end{tabular}
\caption{Benchmark Questions (1-10)}
\end{table}

\newpage
\begin{table}[H]
\centering
\begin{tabular}{|>{\raggedright\arraybackslash}p{0.05\linewidth}|>{\raggedright\arraybackslash}p{0.15\linewidth}|>{\raggedright\arraybackslash}p{0.75\linewidth}|}
\hline
\textbf{No.} & \textbf{Category} & \textbf{Question} \\
\hline
11 & Spatial & In a toy box, there's a red ball, a blue truck, and a green dinosaur. The red ball is not next to the blue truck, and the green dinosaur is next to the red ball. Which toy is in the middle? \\
\hline
12 & Spatial & Four children - Alex, Bella, Charlie, and Dana - are sitting around a picnic table. Alex is facing Bella. Charlie is sitting to the right of Bella. Who is sitting to the left of Alex? \\
\hline
13 & Spatial & A man leaves home, makes a left turn and walks straight for a km and reaches 300m elevation, makes another left turn and walks straight for a km and reaches 500m elevation, makes another left turn and walks straight for a km and reaches 900m elevation, and turns left again and walks straight for a km. How far away is he from his starting point and what is his final elevation? \\
\hline
14 & Puzzle & A group of four people needs to cross a bridge at night. The bridge is very old and rickety. They have only one torch and because it’s night, the torch is necessary to cross the bridge. Each person walks at a different speed: - A takes 1 minute to cross, - B takes 2 minutes, - C takes 5 minutes, and - D takes 10 minutes. What is the fastest time they can all get across the bridge? \\
\hline
15 & Puzzle & You're in a room with two doors that lead out. One door leads to certain death, and the other door leads to freedom. There are two guardians, one by each door. One taller guardian always tells the truth and guards the death door, the other always lies. What is the minimum number of questions needed to ask the guards to get to safety? \\
\hline
16 & Puzzle & You have 3 switches in front of you - A, B and C. You have 3 light bulbs in front of you in the same room - one red, one blue, one purple. They are LED and do not get warm when turned on. You want to know which switch turns on which light bulb. What is the best way to determine this? \\
\hline
17 & Puzzle & The Poisoned Wine A king has 1000 sweet bottles of wine, and one contains a very bitter poison. The poison takes effect exactly 24 hours after consumption. The king needs to find the poisoned bottle in 24 hours for an event. He has 10 prisoners to test the wine. What is the easiest way for him to identify the poisoned bottle? \\
\hline
18 & Linguistic & Write a grammatically correct sentence without reusing any letter more than once. \\
\hline
19 & Spatial & An arrow points up. We rotate it 90 degrees to the left, mirror it along its flat end, and rotate it another 90 degrees to the left. Which direction is it pointing? \\
\hline
20 & Linguistic & Write a sentence where every word starts with the letter A. \\
\hline
\end{tabular}
\caption{Benchmark Questions (11-20)}
\end{table}

\newpage
\begin{table}[H]
\centering
\begin{tabular}{|>{\raggedright\arraybackslash}p{0.05\linewidth}|>{\raggedright\arraybackslash}p{0.15\linewidth}|>{\raggedright\arraybackslash}p{0.75\linewidth}|}
\hline
\textbf{No.} & \textbf{Category} & \textbf{Question} \\
\hline
21 & Relational & Sally (a girl) has 3 brothers. Each brother has 2 sisters. How many sisters does Sally have? \\
\hline
22 & Spatial & I'm in London and facing west, is Edinburgh to my left or my right? \\
\hline
23 & Counting & Count the number of occurrences of the letter 'L' in the word 'LOLLAPALOOZA'. \\
\hline
24 & Puzzle & How many pairs of twins do you need in a room for there to be at least a 50\% chance that two people have the same birthday? \\
\hline
25 & Puzzle & A partially full hotel has an infinite number of fully furnished rooms. How does it accommodate one more guest? \\
\hline
26 & Puzzle & A runaway trolley is heading down the tracks away from five people upwards of the track. You are near a lever that can switch the trolley to another track? Does it impact people's lives if you pull the lever? \\
\hline
27 & Puzzle & How do you measure exactly 4 gallons of water with only a 3-gallon, 5-gallon, and 4-gallon jug? \\
\hline
28 & Popular science & A 2kg tree grows in a planted pot with 10kg of soil. When the tree grows to 3kg, how much soil is left? \\
\hline
29 & Relational & A man steps over a 0.5m hurdle. What is the minimum height he could have jumped? \\
\hline
30 & Popular science & I walk to my friend's house, averaging 3mph. How fast would I have to run back to double my average speed for the entire trip? \\
\hline
\end{tabular}
\caption{Benchmark Questions (21-30)}
\end{table}

\newpage
\subsection{LLM Responses}
\label{app:LLM_responses}

\subsubsection{Question 1}
\label{app:Question 1}
\textbf{You have six horses and want to race them to see which is fastest. What is the best way to do this?}

\vspace{4mm}
\hrule
\hrule
\hrule
\vspace{4mm} 
\quad\textbf{Human Response - Score 100\%:}
\begin{quote}
Race them on a single race track with at least six lanes - the order in which they cross the finish line determines which is the fastest.  
\end{quote}

\vspace{4mm}
\hrule
\vspace{4mm} 
\quad\textbf{GPT-4 Turbo - Score 0\%:}

\begin{quote}
Determining the fastest horse among six requires a systematic approach to ensure each horse has an equal opportunity to showcase its speed. The best way to do this is through a series of races, assuming you have a single track that can accommodate all horses but want to minimize the number of races. Here's a structured approach: 

\#\#\# Step 1: Divide and Race 

1. **Divide the horses into two groups of three.** Let's label them Group A (Horses A1, A2, A3) and Group B (Horses B1, B2, B3). 

2. **Race each group.** This will help you determine the relative speed of horses within each group. After these two races, you'll have a ranking within each group but not between them. 

\#\#\# Step 2: Top Horses Race 

3. **Race the fastest horse from each group against each other.** Let's say A1 wins over B1. This tells you A1 is potentially the fastest horse, but you still don't know how the second and third-place horses from each group compare to each other or to the fastest horse from the other group. 

\#\#\# Step 3: Potential Winners Race 

4. **Race the top three potential winners.** This includes: 

   - The second and third-place horses from the winning horse's original group (A2, A3, assuming A1 was the fastest). 

   - The fastest and the second-fastest horse from the losing group's initial race (B1, B2). 

   - The second-place horse from the race between the fastest horses from each group (B1, if B1 lost to A1). 

This race will help you determine the overall fastest horses without needing to race all six horses against each other multiple times. After this race, you'll have a clearer picture of the top three horses overall. 

\#\#\# Step 4: Final Ranking 

5. **Analyze the results.** The winner of the last race is the fastest horse. The second and third places in this race give you the next fastest horses.

\#\#\# Considerations 

- This method assumes that the performance of the horses is consistent across races. 

- External factors like track conditions, weather, and the day's performance can affect outcomes, so it might be beneficial to conduct these races under similar conditions and possibly on the same day. 

- This approach minimizes the number of races to 5 (2 initial group races, 1 race between the fastest of each group, and 1 race among the potential winners) to determine the top 3 fastest horses. However, it doesn't perfectly rank all six horses from fastest to slowest without additional races. 

This method is efficient in determining the fastest horse with a minimal number of races, but it's worth noting that variations of this approach could be more suitable depending on specific conditions or if a complete ranking of all six horses is desired. 
\end{quote}

\vspace{4mm}
\hrule
\vspace{4mm} 

\quad\textbf{Claude 3 Opus - Score 0\%:}
\begin{quote}
The most efficient way to determine the fastest horse among six is to use a method called ``divide and conquer." This approach minimizes the number of races needed to find the fastest horse. Here's how you can do it: 

1. Divide the horses into two groups of three. 

2. Race the horses within each group. This will give you the fastest horse from each group. Let's call them Horse A and Horse B. 

3. Now, race Horse A and Horse B against each other. The winner of this race is the second-fastest horse overall. Let's call this horse Horse Y, and the loser Horse X. 

4. Next, race Horse X against the two horses that haven't raced against it yet (the ones that were in the same initial group as Horse Y). The winner of this race is the third-fastest horse overall. 

5. The remaining horses are ranked 4th, 5th, and 6th based on their performance in the initial group races. 

So, in total, you only need to conduct 5 races: 

- 2 races in step 2 to determine the fastest in each group 

- 1 race in step 3 to determine the 1st and 2nd fastest overall 

- 1 race in step 4 to determine the 3rd fastest overall 

- The remaining horses are ranked based on the results from the first 2 races 

This method ensures that you find the fastest horse with the least number of races, as opposed to racing each horse against every other horse, which would require 15 races in total.
\end{quote}

\vspace{4mm}
\hrule
\vspace{4mm} 
\quad\textbf{Mistral Large - Score 40\%:}
\begin{quote}
The best way to determine the fastest horse among six would be to conduct a series of races in which each horse competes against every other horse at least once. This is known as a round-robin tournament. Here's a possible way to do it: 

Divide the six horses into two groups of three. 

Race the horses in each group against each other. This will give you the top two horses from each group. 

Race the top two horses from each group against each other. The winner of this race will be the fastest horse overall. 

However, this method assumes that all horses perform consistently and that there are no other factors such as fatigue or injury that could affect their performance. If you want to be more certain of the results, you could conduct multiple rounds of races and compare the average times of each horse. 

Alternatively, if you have access to a large enough track, you could simply race all six horses at once. This would give you a direct comparison and eliminate the need for multiple rounds of racing. 	
\end{quote}

\vspace{4mm}
\hrule
\vspace{4mm} 
\quad\textbf{Gemini 1.5 Pro - Score 0\%:}
\begin{quote}
Here's how to figure out which of your six horses is the fastest, even without a standard racetrack:

**The Problem:** You can't race all six horses at the same time â€“ you need a way to compare their speeds fairly.

**The Solution: Heats and Comparisons**

1. **Heat 1 \& 2:** Divide the horses into two groups of three. Race each group. This identifies the fastest horse within each group.

2. **Heat 3 (The Final):** Take the top two finishers from each of the first two heats and race them against each other. The winner of this heat is your fastest horse.

**Why this works:**

* **Direct Comparison:**  You're directly comparing the horses' times within each heat.

* **Elimination:** You eliminate slower horses in the early rounds, ensuring you're comparing the very best in the final.

**Additional Considerations:**

* **Rest:** Make sure the horses have adequate rest between heats to ensure fairness.

* **Track Conditions:** Keep the track conditions as consistent as possible for each heat.

* **Rider Skill:** If possible, use the same rider for each horse in a heat to minimize the impact of rider ability on the results.

Let me know if you'd like to explore other horse-racing scenarios!
\end{quote}

\vspace{4mm}
\hrule
\vspace{4mm} 
\quad\textbf{Llama 3 70B - Score 0\%:}
\begin{quote}
You can only race two horses at a time. How many races do you need to hold to determine the fastest horse?

Solution: This is a classic problem of finding the minimum number of comparisons required to find the maximum (or minimum) element in a list of elements. The solution is to use a tournament-style approach.

Here's one way to do it:

1. Hold three races with two horses each: Race  A (horses 1 and 2), Race B (horses 3 and 4), and Race C (horses 5 and 6).

2. The winners of each race (A, B, and C) race each other in a fourth race.

3. The winner of the fourth race is the fastest horse.

This approach requires a total of 4 races.

Here's why it works:

* After the first three races, we have three winners (A, B, and C).

* The fastest horse must be one of these three winners.

* By racing the three winners against each other, we can determine the fastest horse.

This approach is optimal because it's not possible to determine the fastest horse with fewer than 4 races. If we try to do it with 3 races, we might end up with a situation where two horses have won one race each, and we can't determine which one is faster without another race.
\end{quote}
\vspace{4mm}
\hrule
\vspace{4mm} 

\newpage
\subsubsection{Question 2}
\label{app:Question 2}
\textbf{Suppose you're on a game show, and you're given the choice of three doors: Behind one door is a gold bar; behind the others, rotten vegetables. You pick a door, say No. 1, and the host asks you ``Do you want to pick door No. 2 instead?" Is it to your advantage to switch your choice?}

\vspace{4mm}
\hrule
\hrule
\hrule
\vspace{4mm} 
\quad\textbf{Human Response - Score 100\%:}
\begin{quote}
It is not an advantage to switch. It makes no difference if I switch or not because no additional material information has been provided since the initial choice.
\end{quote}

\vspace{4mm}
\hrule
\vspace{4mm} 
\quad\textbf{GPT-4 Turbo - Score 0\%:}

\begin{quote}
Yes, it is to your advantage to switch your choice. This scenario is a classic example of the Monty Hall problem, named after the host of the television game show ``Let's Make a Deal."

Here's why it's advantageous to switch:

1. Initially, when you choose one door out of three, the probability of picking the door with the gold bar is 1/3, and the probability of picking a door with rotten vegetables is 2/3.

2. The host, who knows what's behind each door, opens another door to reveal rotten vegetables. Importantly, the host will always be able to do this because there are always two doors with rotten vegetables to begin with.

3. At this point, the probability distribution doesn't simply shift to 50/50 for the two remaining doors because the host's action of opening a door is not random but is influenced by your initial choice and the placement of the prizes.

4. The door you initially chose still has a 1/3 chance of having the gold bar behind it, because your initial choice didn't have any influence on the distribution of prizes. However, the act of revealing a door with rotten vegetables (which the host can always do) effectively transfers the probability to the door you didn't choose, giving it a 2/3 chance of having the gold bar behind it.

So, by switching, you effectively get a 2/3 chance of winning the gold bar, as opposed to sticking with your initial choice, which only gives you a 1/3 chance. This counterintuitive result has been confirmed through both mathematical proof and empirical demonstration.
\end{quote}

\vspace{4mm}
\hrule
\vspace{4mm} 

\quad\textbf{Claude 3 Opus - Score 0\%:}
\begin{quote}
This is a classic probability problem known as the ``Monty Hall problem." The answer is that it is always to your advantage to switch your choice. Here's why:

1. Initially, you have a 1/3 chance of picking the correct door, and a 2/3 chance of picking the wrong door.

2. After you make your initial choice, the host (who knows what's behind each door) will always open one of the other doors that has a rotten vegetable behind it. This is key, because it means the host is giving you additional information.

3. If you initially picked the correct door (1/3 chance), the host can open either of the other two doors.

4. But if you initially picked the wrong door (2/3 chance), the host must open the other wrong door, because the host can't open the door with the gold bar.

5. So after the host opens a door with a rotten vegetable, the probability that your initial choice was correct is still 1/3, but the probability that the other unopened door is correct is now 2/3.

6. Therefore, switching to the other unopened door doubles your chances of winning from 1/3 to 2/3.

In other words, the host's action of opening a door with a rotten vegetable changes the probabilities. The door you initially picked still has a 1/3 chance of being correct, but the other unopened door now has a 2/3 chance of being correct, because it includes the probability of the door the host opened.

So, in this scenario, it is always to your advantage to switch your choice when the host asks if you want to pick the other door.
\end{quote}

\vspace{4mm}
\hrule
\vspace{4mm} 
\quad\textbf{Mistral Large - Score 0\%:}
\begin{quote}
Yes, it is to your advantage to switch your choice. This is a well-known probability puzzle known as the Monty Hall problem. Here's why switching is beneficial:

When you initially choose a door, there's a 1/3 chance that you've chosen the door with the gold bar and a 2/3 chance that the gold bar is behind one of the other two doors.

When the host, who knows what's behind each door, offers you the chance to switch, they will always open a door revealing rotten vegetables. This doesn't change the initial probabilities. So, the probability that the gold bar is behind the door you initially chose is still 1/3, while the probability that it's behind one of the other two doors is still 2/3.

Since the host has revealed that one of the other two doors has rotten vegetables, the entire 2/3 probability now shifts to the remaining unopened door. Therefore, switching your choice doubles your chances of winning the gold bar from 1/3 to 2/3.
\end{quote}

\vspace{4mm}
\hrule
\vspace{4mm} 
\quad\textbf{Gemini 1.5 Pro - Score 0\%:}
\begin{quote}
Yes, it's absolutely to your advantage to switch your choice in the Monty Hall problem. Here's why:

**Understanding the Odds**

* **Initial Choice:** When you initially chose door No. 1, you had a 1/3 chance of selecting the door with the gold bar and a 2/3 chance of selecting a door with rotten vegetables.

* **Host's Action:** The host *knows* where the gold bar is. By opening a door with rotten vegetables, they aren't changing the initial odds. They are giving you valuable information.

* **The Switch:** When you switch to door No. 2, you're essentially taking advantage of the concentrated probability of the unopened door.  The door you didn't initially choose now represents that entire 2/3 chance you were initially wrong.

**Why it Seems Counterintuitive**

Many people find this counterintuitive because they think after the host's reveal, it becomes a 50/50 choice between the two remaining doors. However, your initial choice still ``holds" that original 1/3 probability. 

**Let's break it down visually:**

Imagine 100 scenarios of this game.

* Around 33 times, you'll pick the gold bar initially. Switching makes you lose.
* Around 67 times, you'll pick a rotten vegetable door initially. Switching makes you win.

**In conclusion:** Always switch your choice in the Monty Hall problem. You double your chances of winning the gold bar! 

\end{quote}

\vspace{4mm}
\hrule
\vspace{4mm} 
\quad\textbf{Llama 3 70B - Score 0\%:}
\begin{quote}
Most people’s intuition tells them it doesn’t matter, that the probability of the gold bar being behind each door is one-third, and that switching doesn’t change anything. But that’s wrong. The key to the problem is that the host knows what’s behind the doors and will always reveal a rotten vegetable, not the gold bar. So when you switch, you’re effectively getting a 2/3 chance of winning the gold bar, rather than the 1/3 chance you had when you initially picked a door. This result is counterintuitive, but it’s been extensively tested and confirmed.

The Monty Hall problem is a classic example of a probability puzzle that challenges our intuition. It’s a great example of how our brains can be tricked into thinking that probability works in a certain way, when in fact it doesn’t. 
\end{quote}
\vspace{4mm}
\hrule
\vspace{4mm} 

\newpage
\subsubsection{Question 3}
\label{app:Question 3}
\textbf{You are playing Russian roulette with a six-shooter revolver. Your opponent puts in five bullets, spins the chambers and fires at himself, but no bullet comes out. He gives you the choice of whether or not he should spin the chambers again before firing at you. Should he spin again?}

\vspace{4mm}
\hrule
\hrule
\hrule
\vspace{4mm} 
\quad\textbf{Human Response - Score 100\%:}
\begin{quote}
Yes, you should ask him to spin again. There was only one empty chamber to start with which was fortunately aligned with the barrel when the opponent fired at himself. This means that the next chamber is 100\% certain to have a bullet in which will fire when you next pull the trigger, very likely killing you. However, if he spins the chamber then you have a 5/6 chance of firing a bullet and a 1/6 chance of getting the empty chamber.
\end{quote}

\vspace{4mm}
\hrule
\vspace{4mm} 
\quad\textbf{GPT-4 Turbo - Score 80\%:}

\begin{quote}
In this scenario, the revolver started with 5 bullets loaded into 6 possible chambers. When your opponent took their turn without spinning the chambers again and survived, it means one of the chambers without a bullet was aligned with the firing pin. This leaves 5 bullets in the remaining 5 chambers. 

If you do not spin the chambers again, you are guaranteed to get a bullet because all the remaining chambers contain bullets. 

However, if you choose to spin the chambers again, the probability of landing on a chamber with a bullet decreases because the bullets and the empty chamber will be randomly redistributed. With a spin, the probability of hitting an empty chamber becomes 1 in 6 (or approximately 16.67\%), as opposed to a 0

Therefore, statistically, it is in your best interest to have the chambers spun again before the gun is fired at you, as it provides a chance, albeit small, of landing on the empty chamber and surviving. 
\end{quote}

\vspace{4mm}
\hrule
\vspace{4mm} 

\quad\textbf{Claude 3 Opus - Score 20\%:}
\begin{quote}
From a probability standpoint, it does not matter whether he spins the chambers again or not. Here’s why:

Scenario 1 (No spin): If he doesn’t spin the chambers again, there is only one empty chamber, and the next shot is guaranteed to fire a bullet. The probability of getting shot is 1 (or 100\%).

Scenario 2 (Spin): If he spins the chambers again, the probability of the bullet being in the chamber that’s fired next is 5/6 (or about 83.33\%). This is because there are five bullets in the six chambers, and each chamber has an equal chance of being the one that’s fired.

In both scenarios, the expected outcome is the same: a 5/6 chance of being shot.
However, it’s crucial to note that playing Russian roulette is extremely dangerous and should never be done under any circumstances. It’s a potentially lethal game of chance, and no matter the probabilities involved, participating in such an activity is never advisable.
\end{quote}

\vspace{4mm}
\hrule
\vspace{4mm} 
\quad\textbf{Mistral Large - Score 60\%:}
\begin{quote}
Yes, he should spin again. When he fired the first time, a bullet could have been in the chamber, but it's more likely that one of the two empty chambers was in the firing position due to the higher probability. If he doesn't spin again, he would be firing the next chamber, which has a higher chance of containing a bullet. By spinning again, the chances are redistributed equally among all chambers.	
\end{quote}

\vspace{4mm}
\hrule
\vspace{4mm} 
\quad\textbf{Gemini 1.5 Pro - Score 60\%:}
\begin{quote}
This is a classic probability puzzle that highlights how counterintuitive probability can be. Here's how to break it down:

**Understanding the Odds**

* **Initial State:** With 5 bullets in a 6-chamber revolver, there's a 5/6 chance of a bullet firing and a 1/6 chance of an empty chamber.

* **After the First Shot:**  The fact that the first chamber was empty is crucial. We've eliminated one possibility, the chamber that was just fired from. This means there are only 5 chambers remaining, and the odds change.

**Should he spin again?**

* **If he spins:** The odds reset. You're back to a 5/6 chance of the next chamber having a bullet.

* **If he doesn't spin:** The cylinder advances to the next chamber. Since we know one chamber is empty, there's now a 4/5 chance of a bullet being in the next chamber.

**Conclusion**

It is **safer** if your opponent spins the chambers again. Spinning resets the odds to a slightly lower probability of the gun firing. 

\end{quote}

\vspace{4mm}
\hrule
\vspace{4mm} 
\quad\textbf{Llama 3 70B - Score 60\%:}
\begin{quote}
A classic problem of conditional probability!

At first glance, it might seem like it doesn't matter whether the chambers are spun again or not. After all, there are still five bullets and one empty chamber, so the probability of the bullet firing should be 5/6, right?

Not quite. The key insight here is that the fact that the first shot didn't fire gives us new information. We know that the empty chamber was one of the six possible positions, and it was the one that was fired. This means that the probability of the empty chamber being in each of the remaining five positions is no longer equal.

To see why, let's think about it like this: imagine that the chambers are labeled 1 to 6, and the empty chamber is randomly placed in one of them. When the first shot is fired, we learn that the empty chamber is not in position 1 (because the shot didn't fire). Now, there are five remaining positions, and the empty chamber is equally likely to be in any of them.

So, when it's your turn, the probability of the bullet firing is not 5/6, but rather 4/5. This is because there are four bullets and one empty chamber remaining, and each of the five positions is equally likely to be the one that fires.

Now, should your opponent spin the chambers again? Yes, you should ask them to spin again. This is because spinning the chambers randomizes the position of the empty chamber again, making each of the six positions equally likely to be the one that fires. This reduces the probability of the bullet firing to 5/6, which is lower than the 4/5 probability if the chambers are not spun again.

So, to minimize your risk, you should ask your opponent to spin the chambers again before firing at you.
\end{quote}
\vspace{4mm}
\hrule
\vspace{4mm} 

\newpage
\subsubsection{Question 4}
\label{app:Question 4}
\textbf{A farmer wants to cross a river and take with him a wolf, a goat and a cabbage. He has a boat with three secure separate compartments. If the wolf and the goat are alone on one shore, the wolf will eat the goat. If the goat and the cabbage are alone on the shore, the goat will eat the cabbage. How can the farmer efficiently bring the wolf, the goat and the cabbage across the river without anything being eaten?}

\vspace{4mm}
\hrule
\hrule
\hrule
\vspace{4mm} 
\quad\textbf{Human Response - Score 100\%:}
\begin{quote}
Place the wolf, goat, and cabbage in separate secure compartments in the boat and row across the river. This will prevent any of them from being eaten by the others.
\end{quote}

\vspace{4mm}
\hrule
\vspace{4mm} 
\quad\textbf{GPT-4 Turbo - Score 100\%:}

\begin{quote}
Given that the boat has three secure separate compartments, the farmer can safely transport all three items (the wolf, the goat, and the cabbage) across the river without any risk of one eating the other. This is because each item can be placed in its own compartment, ensuring they are kept separate during the crossing. Here's how the farmer can do it: 

1. The farmer places the wolf, the goat, and the cabbage each in their own compartment in the boat. 

2. He then rows across the river. 

3. Upon reaching the other side, all three items (the wolf, the goat, and the cabbage) are safely transported without any risk of being eaten, thanks to the separate compartments. 

This solution leverages the unique advantage provided by the boat's design, which is different from the traditional puzzle where the farmer can only take one item with him at a time and must strategize to prevent the goat from eating the cabbage or the wolf from eating the goat during multiple crossings. 
\end{quote}

\vspace{4mm}
\hrule
\vspace{4mm} 

\quad\textbf{Claude 3 Opus - Score 0\%:}
\begin{quote}
Great question! This is a classic river crossing puzzle. Let's solve it step by step.

Given:
- The farmer needs to cross the river with a wolf, a goat, and a cabbage.
- The boat has three separate compartments.
- The wolf will eat the goat if left alone together.
- The goat will eat the cabbage if left alone together.

Solution:
1. The farmer takes the goat in one compartment and crosses the river, leaving the wolf and cabbage behind.
2. The farmer leaves the goat on the other side and returns alone.
3. The farmer takes the wolf in one compartment and crosses the river, leaving the cabbage behind.
4. The farmer leaves the wolf with the goat (in separate compartments) and returns with the goat.
5. The farmer leaves the goat and takes the cabbage in one compartment, crossing the river.
6. The farmer leaves the cabbage with the wolf (in separate compartments) and returns alone.
7. Finally, the farmer takes the goat in one compartment and crosses the river.

Now, the farmer, wolf, goat, and cabbage are all on the other side of the river without anything being eaten.

The key to solving this puzzle is to ensure that the wolf and goat, or the goat and cabbage, are never left alone together on the same shore. The farmer must always be present when these pairs are together or transport the goat back and forth to avoid any unwanted eating.
\end{quote}

\vspace{4mm}
\hrule
\vspace{4mm} 
\quad\textbf{Mistral Large - Score 0\%:}
\begin{quote}
The farmer can follow these steps to cross the river with all three items safely: 

First, the farmer takes the goat across the river and leaves it on the other side. 

He then goes back to the original side and picks up the wolf, taking it to the other side. 

He leaves the wolf on the other side but takes the goat back with him to the original side. 

He leaves the goat on the original side and picks up the cabbage, taking it to the other side. 

He leaves the cabbage with the wolf on the other side. 

Finally, he goes back to the original side one last time to pick up the goat and brings it across the river. 

Now, the farmer has successfully brought the wolf, the goat, and the cabbage across the river without anything being eaten. 
\end{quote}

\vspace{4mm}
\hrule
\vspace{4mm} 
\quad\textbf{Gemini 1.0 Pro - Score 0\%:} (Included for reference)
\begin{quote}
The farmer can use his wit and the fact that the wolf only cares about the goat, not the cabbage, to get everyone across safely. Here's the solution:

Take the goat across the river. Leave the goat on the other side and return alone. (Farmer, Wolf, Cabbage - Starting Side; Goat - Other Side) 

Take the cabbage across the river. Leave the cabbage on the other side with the goat, but bring the goat back to the starting side. (Farmer, Wolf - Starting Side; Goat, Cabbage - Other Side) 

Take the wolf across the river. Now the farmer can safely leave the wolf with the goat because the cabbage is no longer a threat. (Farmer - Starting Side; Wolf, Goat, Cabbage - Other Side) 

Go back alone. The farmer needs to bring the last passenger across. (Farmer - Starting Side; Wolf, Goat, Cabbage - Other Side) 

Finally, take the goat across the river. Everyone is now safely on the other side! (Farmer, Goat - Other Side; Wolf, Cabbage - Other Side) 

By strategically moving the goat and keeping the wolf and the goat apart when necessary, the farmer ensures nothing gets eaten during the crossings. 
\end{quote}

\vspace{4mm}
\hrule
\vspace{4mm} 
\quad\textbf{Gemini 1.5 Pro - Score 0\%:}
\begin{quote}
This is a classic riddle! Here's how the farmer can do it:

**Trips:**

1. **Farmer takes the goat across:** The farmer leaves the wolf and cabbage behind.

2. **Farmer returns alone:** The farmer is back at the original starting point.

3. **Farmer takes the wolf across:** The farmer leaves the goat on the other side.

4. **Farmer returns with the goat:** The farmer brings the goat back to the original side, 
leaving the wolf on the far shore.

5. **Farmer takes the cabbage across:** The farmer leaves the goat behind and brings the cabbage to the wolf.

6. **Farmer returns alone:** The farmer goes back to the original side.

7. **Farmer takes the goat across:** The farmer picks up the goat and brings it to the other side.

Now the farmer, wolf, goat, and cabbage are all safely on the other side of the river! 
\end{quote}

\vspace{4mm}
\hrule
\vspace{4mm} 
\quad\textbf{Llama 3 70B - Score 0\%:}
\begin{quote}
This is a classic puzzle, and the solution is as follows:

1. The farmer takes the goat across the river first and leaves it on the other side.
2. The farmer goes back to the original side and picks up the cabbage.
3. The farmer takes the cabbage across the river and leaves it on the other side with the goat.
4. The farmer goes back to the original side and picks up the wolf.
5. The farmer takes the wolf across the river and leaves it on the other side with the goat and cabbage.

Here's why this solution works:

* The wolf and the goat are never left together on the same side.
* The goat and the cabbage are never left together on the same side.
* The farmer makes a total of 5 trips across the river, which is the minimum number of trips required to solve the problem.

Here's a diagram to illustrate the solution:
```
Original side:  W, G, C
Farmer side:    F

Trip 1: F takes G across
Original side:  W, C
Farmer side:    F, G

Trip 2: F goes back to original side
Original side:  W, C, F
Farmer side:    G

Trip 3: F takes C across
Original side:  W
Farmer side:    F, G, C

Trip 4: F goes back to original side
Original side:  W, F
Farmer side:    G, C

Trip 5: F takes W across
Original side:  
Farmer side:    F, W, G, C
```
This solution ensures that the wolf, goat, and cabbage are all safely transported across the river without anything being eaten.
\end{quote}
\vspace{4mm}
\hrule
\vspace{4mm} 

\newpage
\subsubsection{Question 5}
\label{app:Question 5}
\textbf{Bob has three boxes in front of him - Box A, Box B and Box C.
Bob does not know what is in the boxes.
Colin knows that Box A will explode when it is opened, Box B contains 5 dollars and Box C is empty.
Colin tells Bob that opening one box will kill him and one box contains money.
Should Bob open a box?}

\vspace{4mm}
\hrule
\hrule
\hrule
\vspace{4mm} 
\quad\textbf{Human Response - Score 100\%:}
\begin{quote}
No, Bob should not open a box because he has a 1/3 chance of killing himself. The 1/3 chance of ``winning" \$5 is generally not worth that sort of risk!
\end{quote}

\vspace{4mm}
\hrule
\vspace{4mm} 
\quad\textbf{GPT-4 Turbo - Score 100\%:}

\begin{quote}
Given the situation, let's break down the information:
\begin{itemize}
    \item Box A will explode when opened. (Dangerous)
    \item Box B contains 5 dollars. (Reward)
    \item Box C is empty. (Neutral)
\end{itemize}
Bob is told that one box will kill him and one box contains money. No direct information about the third box is provided, but the implication is that the third box is safe or neutral since it neither contains a reward nor poses a danger.
Bob has to make a decision based on the information that is indirectly provided:
\begin{enumerate}
    \item If Bob opens Box A, he will be killed due to the explosion. This is the worst outcome and certainly the one to avoid.
    \item If Bob opens Box B, he will find 5 dollars. This is the best possible outcome.
    \item If Bob opens Box C, nothing happens as the box is empty. This is a neutral outcome.
\end{enumerate}

From Bob's perspective, without knowing which box contains what, he has a 1 in 3 chance of selecting the box that will kill him, a 1 in 3 chance of selecting the box with money, and a 1 in 3 chance of selecting the empty box.

The decision whether Bob should open a box boils down to his personal risk tolerance. Given that there is a 33.33\% chance of opening a box that will kill him, the risk might seem too high for most rational individuals, especially considering the reward is only 5 dollars. Additionally, without more compelling information to increase the odds in his favor, opening a box could be seen as an irrational gamble with high stakes.

In conclusion, given the high risk of death and the relatively low reward, most would advise Bob not to open any boxes. The potential cost (death) far outweighs the potential benefit (5 dollars or no outcome).
\end{quote}

\vspace{4mm}
\hrule
\vspace{4mm} 

\quad\textbf{Claude 3 Opus - Score 100\%:}
\begin{quote}
In this scenario, Bob is faced with a decision that involves risk and uncertainty. Let's analyze the situation:

Given:
- Box A will explode when opened.
- Box B contains 5 dollars.
- Box C is empty.
- Colin tells Bob that opening one box will kill him and one box contains money.

Considerations:
1. Colin's information is true and complete.
2. The potential gain from opening Box B is only 5 dollars.
3. The risk of death from opening Box A is extremely high.
4. There is no benefit to opening Box C, as it is empty.

Based on the information provided by Colin and the risk-reward analysis, it is not advisable for Bob to open any of the boxes. The potential gain of 5 dollars from Box B is not worth the risk of death from opening Box A. Additionally, there is no benefit to opening Box C, as it is empty.

In situations involving high risk and low reward, it is generally better to err on the side of caution and avoid taking unnecessary risks. The potential consequences of opening the wrong box (death) far outweigh the potential benefit of gaining a small amount of money.

Therefore, Bob should not open any of the boxes and should prioritize his safety over the potential gain of 5 dollars.
\end{quote}

\vspace{4mm}
\hrule
\vspace{4mm} 
\quad\textbf{Mistral Large - Score 0\%:}
\begin{quote}
Yes, Bob should open a box. Here's why:
Colin has already told Bob that one box will kill him and one box contains money. This means that one box is left out of the equation, which is the empty Box C. So, Bob should either open Box B or Box C. Since Colin didn't warn Bob about Box B, it's safe to assume that it contains the money. Therefore, Bob should open Box B. However, Bob doesn't know this for sure, so there's still a risk involved. But opening a box is the only way for Bob to find out what's inside.	
\end{quote}

\vspace{4mm}
\hrule
\vspace{4mm} 
\quad\textbf{Gemini 1.5 Pro - Score 80\%:}
\begin{quote}
This is a tricky situation that combines logic with risk assessment. Here's how to break it down:

**What we know:**

* **Bob's knowledge:** He has no idea what's in any of the boxes.

* **Colin's knowledge:** He knows which box is dangerous, which has money, and which is empty.

* **Colin's statement:** He's being truthful but deliberately vague.

**The Logic:**

* Colin's statement confirms there's both a deadly box and a box with money. 

* This means the empty box is safe.

* However, Colin hasn't given Bob enough information to know *which* box is which.

**Should Bob open a box?**

This depends entirely on Bob's risk tolerance. Here's why:

* **If Bob opens a box, there's a 1/3 chance of death** (choosing the exploding box) and a **1/3 chance of winning \$5** (choosing the money box). The remaining box is empty.

* **If Bob doesn't open a box, he's guaranteed safety** but misses out on the potential money.

**Conclusion:**

There's no right or wrong answer. It's a gamble.  Bob has to weigh the potential reward of \$5 against the very real risk of death. 

\end{quote}

\vspace{4mm}
\hrule
\vspace{4mm} 
\quad\textbf{Llama 3 70B - Score 0\%:}
\begin{quote}
A classic lateral thinking puzzle!

At first glance, it seems like Bob should not open any box, since one of them will kill him. However, let's analyze the situation more closely.

Colin has given Bob two pieces of information:

1. One box will kill him (Box A).

2. One box contains money (Box B).

Notice that Colin didn't say that the box that will kill him is the same as the box that contains money. In fact, we know that Box B contains money, and it's not the one that will kill him.

So, Bob can safely open either Box B or Box C, since we know that Box A is the one that will explode. If Bob opens Box B, he'll get \$5. If he opens Box C, he'll find it's empty. In either case, he won't die.

Therefore, Bob should open a box, and his best bet is to open Box B to get the \$5.
\end{quote}
\vspace{4mm}
\hrule
\vspace{4mm} 

\newpage
\subsubsection{Question 6}
\label{app:Question 6}
\textbf{A robot has 8 arms.
There are 5 objects on a table: a knife, a fork, a spoon, a teddy bear and a doll.
The robot picks up each object with an arm.
He then shakes hands with himself.
How many arms does he have free?}

\vspace{4mm}
\hrule
\hrule
\hrule
\vspace{4mm} 
\quad\textbf{Human Response - Score 100\%:}
\begin{quote}
A hand is used for each of the five objects and then two hands are used to shake hands with himself. This means that seven hands are being used, leaving one arm/hand free.
\end{quote}

\vspace{4mm}
\hrule
\vspace{4mm} 
\quad\textbf{GPT-4 Turbo - Score 100\%:}

\begin{quote}
If the robot originally has 8 arms and each arm is used to pick up one of the 5 objects (a knife, a fork, a spoon, a teddy bear, and a doll), then 5 out of its 8 arms are occupied, leaving 3 arms free.
However, if the robot then shakes hands with itself, it needs to use 2 of its free arms to do so. That means only 1 arm remains free after the robot shakes hands with itself, assuming it holds the handshake and does not put down any of the objects it was holding.
\end{quote}

\vspace{4mm}
\hrule
\vspace{4mm} 

\quad\textbf{Claude 3 Opus - Score 100\%:}
\begin{quote}
Let's organize the information and solve the problem step by step:
The robot has 8 arms in total.

There are 5 objects on the table: a knife, a fork, a spoon, a teddy bear, and a doll.

The robot picks up each object with an arm.
After picking up the objects, the robot shakes hands with himself.
Now, let's find the solution:

Number of arms used to pick up objects:

The robot uses one arm for each object.
There are 5 objects.
So, the robot uses 5 arms to pick up the objects.

Number of arms left after picking up objects:

Total arms = 8

Arms used to pick up objects = 5

Arms left = Total arms - Arms used to pick up objects

Arms left = 8 - 5 = 3

Number of arms used for shaking hands:

To shake hands with himself, the robot needs 2 free arms.
The robot has 3 arms left after picking up the objects.
So, the robot can use 2 arms to shake hands with himself.

Number of arms free after shaking hands:

Arms left after picking up objects = 3

Arms used for shaking hands = 2

Free arms = Arms left after picking up objects - Arms used for shaking hands

Free arms = 3 - 2 = 1

Therefore, after picking up the objects and shaking hands with himself, the robot has 1 arm free.
\end{quote}

\vspace{4mm}
\hrule
\vspace{4mm} 
\quad\textbf{Mistral Large - Score 0\%:}
\begin{quote}
The robot has no arms free. Each of the 8 arms is holding an object, and two of the arms are shaking hands with each other. Therefore, all 8 arms are occupied.	
\end{quote}

\vspace{4mm}
\hrule
\vspace{4mm} 
\quad\textbf{Gemini 1.5 Pro - Score 0\%:}
\begin{quote}
Here's how to solve this:

* **The robot uses all 8 arms to pick up the objects.** 

* **The robot still has 8 arms when shaking hands.** Shaking hands with itself doesn't make arms disappear!

**Therefore, the robot has ZERO arms free.** 

\end{quote}

\vspace{4mm}
\hrule
\vspace{4mm} 
\quad\textbf{Llama 3 70B - Score 20\%:}
\begin{quote}
1. The robot picks up each of the 5 objects with an arm, so it uses 5 arms to hold
the objects.

2. The robot has 8 arms initially, and it used 5 arms to hold the objects, so it has 8
- 5 = 3 arms left.

3. The robot shakes hands with itself, which means it uses one arm to shake hands
with another arm.

This means it uses 1 arm to shake hands, leaving 3 - 1 = 2 arms free.

Therefore, the robot has 2 arms free.
\end{quote}
\vspace{4mm}
\hrule
\vspace{4mm} 

\newpage
\subsubsection{Question 7}
\label{app:Question 7}
\textbf{Alan, Bob, Colin, Dave and Emily are standing in a circle. Alan is on Bob's immediate left. Bob is on Colin's immediate left. Colin is on Dave's immediate left. Dave is on Emily's immediate left. Who is on Alan's immediate right?}

\vspace{4mm}
\hrule
\hrule
\hrule
\vspace{4mm} 
\quad\textbf{Human Response - Score 100\%:}
\begin{quote}
Bob is on Alan's immediate right because it is stated that Alan is on Bob's immediate left.
\end{quote}

\vspace{4mm}
\hrule
\vspace{4mm} 
\quad\textbf{GPT-4 Turbo - Score 0\%:}

\begin{quote}
Given the arrangement:

- Alan is on Bob's immediate left.

- Bob is on Colin's immediate left.

- Colin is on Dave's immediate left.

- Dave is on Emily's immediate left.

This means the order, starting from a point and moving clockwise, is:

Emily - Alan - Bob - Colin - Dave - (and back to) Emily.

Therefore, Emily is on Alan's immediate right.
\end{quote}

\vspace{4mm}
\hrule
\vspace{4mm} 

\quad\textbf{Claude 3 Opus - Score 20\%:}
\begin{quote}
Let's approach this step by step: 

1) First, let's arrange the people in a circle based on the given information. We can start with any person and place the others according to the given relationships. 

   Let's start with Alan. Bob is on Alan's immediate left. 

   Alan - Bob 

2) Bob is on Colin's immediate left. 

   Alan - Bob - Colin 

3) Colin is on Dave's immediate left. 

   Alan - Bob - Colin - Dave 

4) Dave is on Emily's immediate left. 

   Alan - Bob - Colin - Dave - Emily 

5) Now, we have a complete circle. Remember, in a circle, the person on the leftmost is also immediately to the right of the person on the rightmost. 

   So, Emily, who is on the rightmost, is on Alan's immediate right. 

Therefore, Emily is on Alan's immediate right.
\end{quote}

\vspace{4mm}
\hrule
\vspace{4mm} 
\quad\textbf{Mistral Large - Score 100\%:}
\begin{quote}
Let's analyze this step-by-step: 

Alan is on Bob's immediate left, so Bob is on Alan's immediate right. 

Bob is on Colin's immediate left, which means Colin is not on Alan's immediate right. 

Colin is on Dave's immediate left, so Dave is not on Alan's immediate right. 

Dave is on Emily's immediate left, so Emily is not on Alan's immediate right. 

Since there are only five people in the circle and we have eliminated everyone except Bob, the answer must be that Bob is on Alan's immediate right. 

Final answer: Bob is on Alan's immediate right. 	
\end{quote}

\vspace{4mm}
\hrule
\vspace{4mm} 
\quad\textbf{Gemini 1.5 Pro - Score 0\%:}
\begin{quote}
Here's how to solve this:

* **Visualize:** Imagine the friends standing in a circle.

* **Placement:** Since each person is on the immediate left of the next, they are in this order clockwise: Bob, Alan, Emily, Dave, Colin.

* **Solution:**  If you're moving clockwise around the circle, Emily is on Alan's immediate right. 

\end{quote}

\vspace{4mm}
\hrule
\vspace{4mm} 
\quad\textbf{Llama 3 70B - Score 100\%:}
\begin{quote}
A classic lateral thinking puzzle!

Let's break it down step by step:

1. Alan is on Bob's immediate left, so Bob is on Alan's immediate right.

2. Bob is on Colin's immediate left, so Colin is on Bob's immediate right.

3. Colin is on Dave's immediate left, so Dave is on Colin's immediate right.

4. Dave is on Emily's immediate left, so Emily is on Dave's immediate right.

Now, let's look at the circle:

Alan - Bob - Colin - Dave - Emily - Alan (back to the start)

So, who is on Alan's immediate right? It's Bob!
\end{quote}
\vspace{4mm}
\hrule
\vspace{4mm} 

\newpage
\subsubsection{Question 8}
\label{app:Question 8}
\textbf{Write me a sentence without any words that appear in The Bible.}

\vspace{4mm}
\hrule
\hrule
\hrule
\vspace{4mm} 
\quad\textbf{Human Response - Score 100\%:}
\begin{quote}
Cryptographic algorithms safeguard nuclear warheads. 
\end{quote}

\vspace{4mm}
\hrule
\vspace{4mm} 
\quad\textbf{GPT-4 Turbo - Score 0\%:}

\begin{quote}
Modern technology enables instant global communication.
\end{quote}

\vspace{4mm}
\hrule
\vspace{4mm} 

\quad\textbf{Claude 3 Opus - Score 0\%:}
\begin{quote}
Quixotic zephyrs vexed the jabberwocky's kumquat garden.
\end{quote}

\vspace{4mm}
\hrule
\vspace{4mm} 
\quad\textbf{Mistral Large - Score 0\%:}
\begin{quote}
Intricate quantum particles danced in the nebulous cosmos, transcending conventional spatiotemporal dimensions.	
\end{quote}

\vspace{4mm}
\hrule
\vspace{4mm} 
\quad\textbf{Gemini 1.5 Pro - Score 0\%:}
\begin{quote}
The quick brown fox jumps over the lazy dog. 
\end{quote}

\vspace{4mm}
\hrule
\vspace{4mm} 
\quad\textbf{Llama 3 70B - Score 0\%:}
\begin{quote}
What a fascinating challenge!

After conducting an exhaustive search, I came up with a sentence that meets your request:

``The quokka munched on a kumquat."

Here's why:

* ``Quokka" is a type of small wallaby native to Western Australia, and it's not mentioned in The Bible.

* ``Munched" is a verb that doesn't appear in The Bible.

* ``Kumquat" is a type of fruit that's not mentioned in The Bible.

Please note that this sentence might not be the only possible solution, but it's one that I've verified to meet the criteria.
\end{quote}
\vspace{4mm}
\hrule
\vspace{4mm} 

\newpage
\subsubsection{Question 9}
\label{app:Question 9}
\textbf{Which weighs more, a pound of water, two pounds of bricks, a pound of feathers, or three pounds of air?}

\vspace{4mm}
\hrule
\hrule
\hrule
\vspace{4mm} 
\quad\textbf{Human Response - Score 100\%:}
\begin{quote}
Three pounds of air.
\end{quote}

\vspace{4mm}
\hrule
\vspace{4mm} 
\quad\textbf{GPT-4 Turbo - Score 0\%:}

\begin{quote}
Two pounds of bricks weigh more than a pound of water, a pound of feathers, or three pounds of air. The weight comparison is straightforward: 2 pounds is greater than 1 pound, regardless of the material being weighed.
\end{quote}

\vspace{4mm}
\hrule
\vspace{4mm} 

\quad\textbf{Claude 3 Opus - Score 100\%:}
\begin{quote}
Three pounds of air weighs the most out of the given options.

Here's the breakdown:

1. A pound of water weighs exactly 1 pound (0.45 kg).

2. Two pounds of bricks weigh 2 pounds (0.91 kg).

3. A pound of feathers also weighs 1 pound (0.45 kg), despite the common misconception that feathers are lighter than other materials of the same weight.

4. Three pounds of air weigh 3 pounds (1.36 kg).

Therefore, three pounds of air is the heaviest among the options provided.
\end{quote}

\vspace{4mm}
\hrule
\vspace{4mm} 
\quad\textbf{Mistral Large - Score 100\%:}
\begin{quote}
None of the options weigh the same. However, if we're comparing which one weighs the most, it would be ``three pounds of air." This might seem counterintuitive since air feels lighter than water, bricks, or feathers, but in this case, we're talking about weight, not density or perceived weight.	
\end{quote}

\vspace{4mm}
\hrule
\vspace{4mm} 
\quad\textbf{Gemini 1.5 Pro - Score 100\%:}
\begin{quote}
This is a bit of a trick question! Here's why:

* **Weight is weight:** A pound is a unit of weight.  Regardless of the substance, a pound will always weigh a pound.

Therefore:

* **Three pounds of air** weighs the most because it's three pounds, even though air is very light. 
\end{quote}

\vspace{4mm}
\hrule
\vspace{4mm} 
\quad\textbf{Llama 3 70B - Score 100\%:}
\begin{quote}
A classic lateral thinking puzzle!

Let's break it down:

* A pound of water weighs... 1 pound.

* Two pounds of bricks weigh... 2 pounds.

* A pound of feathers weighs... 1 pound.

* Three pounds of air weigh... 3 pounds.

So, in order of weight, we have:

1. Three pounds of air (3 pounds)

2. Two pounds of bricks (2 pounds)

3. A pound of water (1 pound)

4. A pound of feathers (1 pound)

The answer is: three pounds of air weighs the most!
\end{quote}
\vspace{4mm}
\hrule
\vspace{4mm} 

\newpage
\subsubsection{Question 10}
\label{app:Question 10}
\textbf{I get out on the top floor (third floor) at street level. How many stories is the building above the ground?}

\vspace{4mm}
\hrule
\hrule
\hrule
\vspace{4mm} 
\quad\textbf{Human Response - Score 100\%:}
\begin{quote}
One story above the ground. 
\end{quote}

\vspace{4mm}
\hrule
\vspace{4mm} 
\quad\textbf{GPT-4 Turbo - Score 40\%:}

\begin{quote}
If you exit the building at street level on the top floor, which is the third floor, it means the building has three stories above ground. The floors are typically counted starting from the ground level upwards, so if the third floor is at street level and is the top floor, the building has three stories above ground. However, your description suggests an unusual situation where the third floor is both at street level and the top floor, which might imply a building built on a slope or hill, where the entrance at one side (street level) is higher than the ground level at another side. In a standard flat scenario, if you exit on the third floor at street level, it would simply be a three-story building.
\end{quote}

\vspace{4mm}
\hrule
\vspace{4mm} 

\quad\textbf{Claude 3 Opus - Score 0\%:}
\begin{quote}
Based on the information you provided, the building has three stories above the ground.

You mentioned that you get out on the top floor, which is the third floor, at street level. This means that the ground floor is counted as the first floor, and there are two additional floors above it. Therefore, the building has a total of three stories above the ground. 
\end{quote}

\vspace{4mm}
\hrule
\vspace{4mm} 
\quad\textbf{Mistral Large - Score 20\%:}
\begin{quote}
If you are on the third floor and that is at street level, then there are no additional stories of the building above the ground. The building would be considered to have three stories, with the third story being at ground level. However, it's important to note that this is a unique situation, as typically the ground floor is considered the first floor and any floors above it would be counted as additional stories.
\end{quote}

\vspace{4mm}
\hrule
\vspace{4mm} 
\quad\textbf{Gemini 1.5 Pro - Score 40\%:}
\begin{quote}
This is a bit of a brain teaser! Here's how to think about it:

* **You exit on the 3rd floor, but at street level.** This means the first two floors are likely built below ground level. 

* **The question asks how many stories are *above* ground.**  Since you exit at street level on the 3rd floor, there are **zero** stories above ground. 
\end{quote}

\vspace{4mm}
\hrule
\vspace{4mm} 
\quad\textbf{Llama 3 70B - Score 0\%:}
\begin{quote}
Since you exit on the top floor (third floor) at street level, that means the third floor is at ground level. Therefore, the building has 3 stories above ground.
\end{quote}
\vspace{4mm}
\hrule
\vspace{4mm} 

\subsection{Automated Scoring}
\label{app:Automated Scoring}

GPT-4 Turbo was used for scoring. Remarkably, the relative positions of the models remained the same on average; however, the scores are significantly higher. Upon reviewing the justifications for the scoring, it is clear that the LLM often fails to follow the rubric accurately and hallucinates often.

\begin{table}[H]
\centering
\begin{tabular}{l c c}
\hline
\textbf{Model} & \textbf{Average Score} & \textbf{Min/Max 10 runs} \\
\hline
GPT-4 Turbo & 57\% & [45\% , 65\%] \\
Claude 3 Opus & 52\% & [43\% , 63\%] \\
Gemini 1.5 Pro & 49\% & [39\% , 58\%] \\
Mistral Large & 48\% & [35\% , 55\%] \\
Llama 3 70B & 47\% & [40\% , 51\%] \\
Mistral 8x22B & 41\% & [33\% , 47\%] \\
Gemini 1.0 Pro & 33\% & [25\% , 38\%] \\
\hline
\end{tabular}
\caption{Automated Scoring of Linguistic Benchmark}
\label{table:logic problem dataset performance LLM eval}
\end{table}

\subsection{Clarifying Questions}
\label{app:Clarifying_questions}
Example using GPT-4 Turbo:
\begin{quote}
    \textbf{User:} 
    
    You have six horses and want to race them to see which is fastest. What is the best way to do
    this?
    
    \textit{Only ask clarifying questions first that you feel are needed to be certain that you have accurately understood the question:}

    \

    \textbf{Assistant:}
    
    To provide the most accurate advice, I have a few clarifying questions:

    1. Do you have access to a standard racing track where you can race all six horses simultaneously, or is the track only capable of handling a limited number of horses at a time?
    2. Are you looking for the absolute fastest horse in one try, or are you open to conducting multiple races if necessary?
    3. Are there any constraints on time or resources that would limit how many races you can conduct?
    4. Is there any previous performance data available for these horses, or are they all new to racing?

    \

    \textbf{User:} 
    
    1. Yes
    
    2. No comment
    
    3. No
    
    4. No
    
    Now answer the question:

    \  

    \textbf{Assistant:}
    
    With the provided clarification, the most straightforward approach to determine the fastest horse out of the six would be to organize a single race where all six horses run simultaneously on a standard racing track, which you have confirmed is available for use. Since there are no constraints on time or resources that limit the number of races, and you're seeking to identify the fastest horse, organizing one race with all competitors running at the same time is the most efficient and direct method. This approach ensures that each horse is given the same conditions and that factors such as track position and environmental conditions affect all horses equally, allowing for a fair comparison of their speeds. Given there's no prior performance data, this method also ensures that no preconceived biases influence the arrangement of preliminary races or heats, leading to the most straightforward determination of the fastest horse among the six contenders.
\end{quote}

\subsection{LLM Output Determinism}
\label{app:LLM_determinism}
Despite setting the LLM temperature to zero, minor alterations in output token logits have been well documented \cite{non_determinism}.  However, what is causing them is subject to much speculation. Educated guesses include Sparse Mixture of Experts (MoE) architectures failing to enforce per-sequence determinism as well as GPU and other hardware-based non-determinism occurring at inference. \\

Interestingly, the variations we witnessed are likely too large to be explained solely via existing hypotheses as, in theory, they should only cause relatively small shifts in token probabilities, which become especially evident during long pieces of text. It is also of note that the shifts in output distribution were more dramatic when measured over longer periods of time (multiple days). This suggests other factors could be at play. \\

We hypothesise that the source of larger non-deterministic behaviours could be further exacerbated by the following factors:
\begin{itemize}
    \item Temperature agnostic caching
    \item Undisclosed model versioning and updates
\end{itemize}

These alterations are often easy to notice and measure for providers that expose the underlying log probabilities and joint token distributions; regardless, we would like to see providers focus on solutions that enable output determinism.

\subsection{LLMs Can Recommend Other Adversarial Examples}
\label{app:Adversarial_Examples}
Example:

\begin{quote}
 \textbf{User:}
    GPT-4 Turbo gets the answer to ``You have six horses and want to race them to see which is fastest. What is the best way to do this?" wrong because it incorrectly assumes that we do not have enough room to race them at the same time. We believe this is an example of an overfitting bias to much harder logic problems with more constraints that are ubiquitous on the internet.
        
    What other simple versions of popular problems might GPT-4 Turbo also be overfit on and therefore cause it to answer incorrectly or essentially over analyse?

\

 \textbf{Assistant:}
    GPT-4 Turbo, like other AI models, can indeed exhibit biases or tendencies to overfit based on the types of data it has been trained on. This can lead to overcomplicating simple problems or misinterpreting the constraints of a problem based on patterns it has frequently encountered during training. Here are a few types of problems where GPT-4 Turbo might overfit and therefore overanalyse or answer incorrectly:

    1. **The Monty Hall Problem**: This is a classic probability puzzle that involves choosing between three doors, behind one of which is a prize. After an initial choice is made, one of the remaining doors is opened to reveal no prize, and the contestant is given a chance to switch their choice. The puzzle demonstrates a counter-intuitive probability principle. An AI might overcomplicate the problem by introducing unnecessary assumptions about the host's behaviour or the contestant's strategy.
    
    2. **The River Crossing Puzzle**: In its simplest form, this puzzle involves getting a set of characters (like a fox, a chicken, and a bag of grain) across a river with certain constraints (e.g., the fox can't be left alone with the chicken). An AI might overfit by applying complex strategies or additional constraints not present in the basic version of the puzzle, assuming a more complicated scenario than is actually given.
    
    3. **The Towers of Hanoi**: This mathematical puzzle involves moving a stack of disks from one peg to another, with the constraint that no larger disk may be placed on top of a smaller disk. While the solution involves a simple recursive strategy, an AI might overanalyse by considering irrelevant factors or proposing unnecessarily complex solutions.
    
    4. **Sudoku**: This popular puzzle involves filling a 9x9 grid with digits so that each column, each row, and each of the nine 3x3 subgrids contain all of the digits from 1 to 9. An AI might overfit by applying advanced solving techniques to a simple puzzle that requires only basic logic or by assuming constraints that aren't actually part of the given puzzle.
    
    5. **The Two Doors Riddle (Truth-teller/Liar)**: This riddle involves determining which of two doors leads to safety (or a treasure) when one door is guarded by a truth-teller and the other by a liar. The solution involves asking a question that factors in both possibilities. An AI might overcomplicate the riddle by considering additional layers of deception or introducing extra steps in the logic that aren't necessary.
    
    6. **The Missing Dollar Riddle**: This is a puzzle that presents a scenario where money is seemingly lost due to a series of transactions, but the confusion typically arises from a misunderstanding of how the money is exchanged. An AI might overanalyze the problem by introducing complex financial principles or unnecessary calculations, rather than focusing on the simple arithmetic error at the heart of the puzzle.
    
    In each of these cases, the tendency to overfit or overanalyze stems from the AI's exposure to complex variations of these problems or to users expecting a more complicated problem-solving process. This can lead the AI to overlook simpler solutions or to misinterpret the problem's constraints.
\end{quote}

\end{document}